\documentclass{article}
\usepackage{spconf,amsmath,graphicx,hyperref}
\usepackage{multirow}
\usepackage{algorithm}
\usepackage{algpseudocode}
\usepackage{enumitem}
\usepackage{amssymb}
\usepackage{booktabs}

\title{SIMANF: Sample Free Learning of Unnormalized Distributions via Simulated Annealing in Normalizing Flows}
\name{Vikas Kanaujia}
\address{Mahindra University\\
         Department of Electrical and Computer Engineering\\
         Hyderabad, Telangana, India}
\begin{document}
\ninept
\maketitle
\begin{abstract}
Efficiently learning and sampling from high dimensional, multimodal unnormalized distributions without target samples remains a challenging problem. Although normalizing flows can generate samples efficiently, training based on the reverse KL divergence using only the unnormalized target density may suffer from mode collapse.  We introduce SIMANF, a sample free framework that integrates simulated annealing with normalizing flows. SIMANF progressively transforms the target distribution from a smooth initial form to the original target distribution and trains the flow sequentially across these stages. By transferring the learned representation between stages, the method promotes mode coverage while progressively capturing finer features of the target distribution. Following annealing, a final refinement stage combines the reverse KL divergence with an importance weighted forward KL objective using samples generated by the flow. SIMANF requires no target samples during training and uses only the unnormalized density. We demonstrate its effectiveness on Many-Well distributions and high dimensional Scalar $\phi^4$ lattice field theory distribution.
\end{abstract}

\section{Introduction}
\label{sec:intro}

Unnormalised probability distributions arise naturally across a broad range of scientific domains, including statistical physics~\cite{Pathria,huang2008statistical}, chemistry~\cite{mcquarrie2008quantum}, biology~\cite{bialek2012biophysics}, applied mathematics~\cite{koller2009probabilistic,PARK20124883}, and Bayesian inference~\cite{WELLING200319}. In many such problems, the distribution is specified through an energy, Hamiltonian, or cost function $H(x)$ as $p(x) \propto e^{-H(x)}$, where the normalization constant is often computationally intractable. Characterizing these systems requires estimating expectations of relevant quantities, which in turn necessitates representative samples. Efficient sampling is therefore a fundamental challenge, particularly for high-dimensional and multimodal distributions. Conventional approaches such as Markov chain Monte Carlo (MCMC)~\cite{geyer2011introduction,hastings1970monte,duane1987hybrid} and molecular dynamics (MD)~\cite{frenkel1957understanding} can struggle in such settings~\cite{neal1993probabilistic,frenkel1957understanding}. Their samples are typically correlated, requiring long trajectories to obtain effectively independent samples, while transitions between well-separated modes can be rare, leading to slow mixing and critical slowing down~\cite{leimkuhler2004simulating}.

Deep generative models offer an alternative approach by learning expressive representations of complex, high-dimensional distributions, thereby addressing some of the limitations of conventional sampling methods. Several classes of generative models have been developed, including variational autoencoders (VAEs)~\cite{kingma2019introduction}, normalizing flows (NFs)~\cite{NF_survey_1}, generative adversarial networks (GANs)~\cite{goodfellow2020generative}, score-based models~\cite{song2020score}, and diffusion models~\cite{ho2020denoising}. Among these, NFs are particularly attractive for unnormalised target distributions because they provide tractable and exact likelihood evaluation~\cite{NF_survey_1}. This property also enables independent Metropolis--Hastings (IMH) correction and importance weighting, providing mechanisms for asymptotically exact sampling and bias correction, respectively~\cite{noe2019boltzmann,midgleyflow}.

NFs, commonly used as Boltzmann Generators (BGs)~\cite{noe2019boltzmann}, can be trained by minimizing the Kullback--Leibler (KL) divergence between the model and target distributions. The choice of KL direction leads to an important trade-off. Forward KL (FKL) promotes mode coverage but can allocate probability mass to low density regions, resulting in high-variance estimates of observables~\cite{midgleyflow,kanaujia2024advnf}. More importantly, FKL training requires samples from the target distribution, typically obtained using MCMC, thereby reintroducing the sampling cost. Reverse KL (RKL), in contrast, can be optimized directly from evaluations of the unnormalised density and therefore does not require target samples. However, its mode seeking behavior can concentrate probability mass around dominant modes while neglecting low-probability modes, leading to mode collapse and biased estimates~\cite{midgleyflow,kanaujia2024advnf,kanaujia2025scorenf}.

This trade-off exposes a key limitation in learning complex unnormalised distributions without target samples. While RKL enables sample free training, its mode seeking behavior can limit coverage of multimodal targets~\cite{kanaujia2024advnf,nicoli2023detecting}. Conversely, obtaining target samples to enable FKL training through conventional samplers can reintroduce the computational bottleneck that generative models seek to avoid. This motivates sample free generative approaches that improve mode coverage while relying only on evaluations of the underlying energy function.

Recent sample free approaches include Flow Annealed Importance Sampling Bootstrap (FAB)~\cite{midgleyflow}, Iterated Denoising Energy Matching (iDEM)~\cite{Akhound-SadeghR24}, and FUND~\cite{kanaujia2026fund}. FAB incorporates annealed importance sampling (AIS) into the NF framework and optimizes an $\alpha$-divergence using AIS-generated samples, but its reliance on computationally intensive transition steps, such as Hamiltonian Monte Carlo (HMC), can make training costly in high dimensions. iDEM trains a diffusion-based sampler using Monte Carlo score estimates, which may exhibit high variance and requires repeated simulation of the reverse stochastic differential equation (SDE). Its lack of an exact likelihood also precludes direct importance weighting or independent Metropolis--Hastings (IMH) correction. FUND instead trains the flow sequentially on reshaped target distributions using previously generated samples, but introduces additional computational cost and sensitivity to hyperparameter choices.

To address these limitations, we introduce \textbf{SIMANF}, a sample free framework for learning unnormalised distributions using normalizing flows and simulated annealing. Rather than directly optimizing the reverse KL divergence for the target distribution, SIMANF constructs a sequence of distributions that gradually approaches the target distribution and trains the flow sequentially along this annealing path. By transferring the learned flow parameters between successive stages, the method facilitates coverage of multiple modes while progressively refining the learned distribution toward the target. After annealing, a refinement stage combines reverse KL with an importance weighted forward KL objective using flow-generated samples. The proposed framework requires no samples from the target distribution during training and relies only on evaluations of its unnormalised density. Our main contributions are:
\begin{enumerate}
    \item We propose \textbf{SIMANF}, a sample free normalizing flow framework that integrates simulated annealing with reverse KL (RKL) training to mitigate mode collapse when learning high-dimensional and multimodal unnormalised distributions.

    \item We formulate a weighted reverse KL objective over a sequence of annealed target distributions, yielding an effective annealing parameter that provides a gradual transition from a smoothed distribution to the target distribution while transferring the learned representation across successive stages.
    
    \item We introduce a post annealing refinement stage that combines reverse KL (RKL) with an importance weighted forward KL (FKL) objective using flow generated samples, improving sampling efficiency without requiring samples from the true target distribution.

    \item We demonstrate the effectiveness of SIMANF on Many-Well distributions and high dimensional scalar $\phi^4$ lattice field theory, showing its ability to learn complex multimodal distributions while enabling efficient sampling.
\end{enumerate}

\begin{figure}
    \centering
    \includegraphics[width=1.0\linewidth]{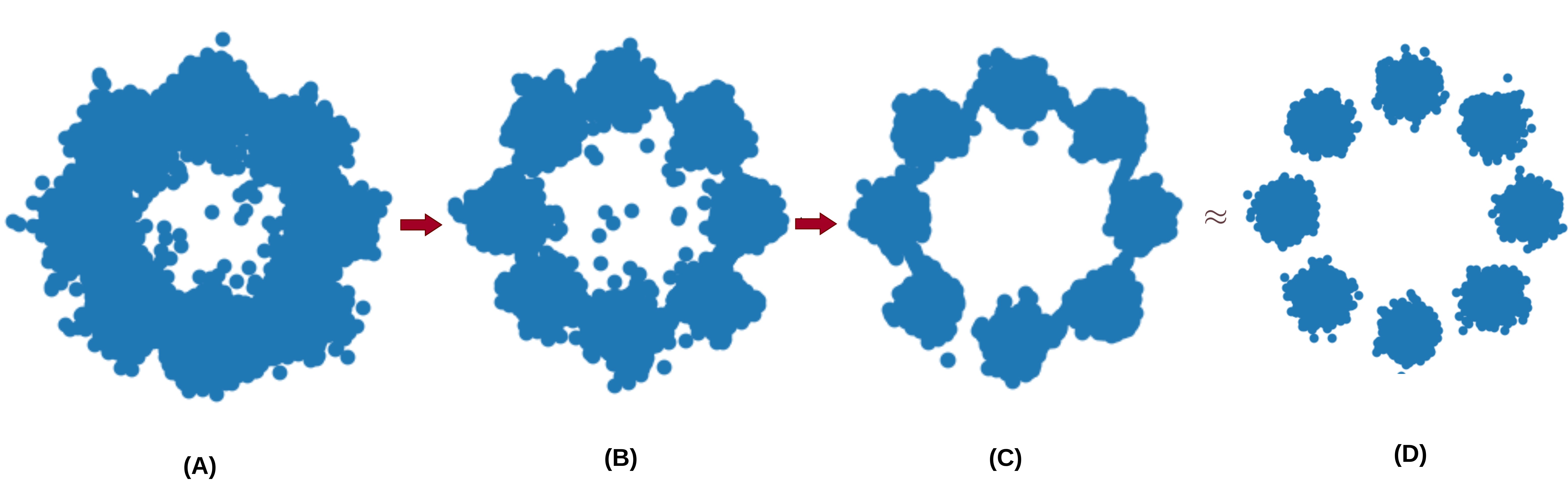}
    \caption{\small{Evolution of generated samples for the MOG-8 distribution across successive annealing stages: (A)--(C) show samples generated at successive stages, while (D) shows samples from the target distribution.}}
    \label{fig1}
\end{figure}

\begin{table*}[ht]
    \centering
    \scalebox{0.95}{
    \begin{tabular}{l|ccc|ccc}
    \toprule
     & \multicolumn{3}{c|}{\textbf{MW-8} $(d =8)$} & \multicolumn{3}{c}{\textbf{MW-16} $(d =16)$} 
    \\\cmidrule(lr){2-4}\cmidrule(lr){5-7}
     & NLL$\downarrow$ & RNLL($\downarrow$) & ESS($\uparrow$) & NLL$\downarrow$ & RNLL($\downarrow$) & ESS($\uparrow$)\\\midrule
    BG (FKL) & 7.60 $\pm$ 0.05 & -31.60 $\pm$ 0.77 & 0.34 $\pm$ 0.02 & 17.14 $\pm$ 0.07 & -48.57 $\pm$ 5.77 & 0.001 $\pm$ 0.004 \\
    BG (RKL) & 76.72 $\pm$ 73.84 & -35.40 $\pm$ 0.01 & \textbf{0.97} $\pm$ \textbf{0.03} & 604.83 $\pm$ 85.65 & -70.80 $\pm$ 0.06 & \textbf{0.83} $\pm$ \textbf{0.06} \\
    FAB & 7.00 $\pm$ 0.02 & -33.87 $\pm$ 0.10 & 0.81 $\pm$ 0.01 & 14.28 $\pm$ 0.01 & -66.53 $\pm$ 0.13 & 0.31 $\pm$ 0.02 \\
    iDEM & 7.64 $\pm$ 0.09 & -34.86 $\pm$ 0.02 & 0.55 $\pm$ 0.03 & 15.32 $\pm$ 0.43 & -70.08 $\pm$ 0.36 & 0.25 $\pm$ 0.03 \\
    FUND (FKL) & 6.97 $\pm$ 0.01 & -34.25 $\pm$ 0.04 & 0.87 $\pm$ 0.02 & 14.04 $\pm$ 0.05 & -67.95 $\pm$ 0.25 & 0.71 $\pm$ 0.04\\
    \midrule
    SIMNF & \textbf{6.97} $\pm$ \textbf{0.002} & -34.00 $\pm$ 0.10 & \textbf{0.91} $\pm$ \textbf{0.004} & \textbf{14.03} $\pm$ \textbf{0.02} & -67.85 $\pm$ 0.08 & \textbf{0.71} $\pm$ \textbf{0.01}\\
    
    \bottomrule
    \end{tabular}
    }
    \vspace{-1mm}
    \caption{Results for the MW-8 and MW-16 distributions. Reported values represent the average and standard error computed over three random seeds for each method.}
    \label{tab:mw_results}
\end{table*}

\begin{table*}[ht]
    \centering
    \scalebox{0.95}{
    \begin{tabular}{l|ccc|ccc}
    \toprule
     & \multicolumn{3}{c|}{$8\times8$} & \multicolumn{3}{c}{$10\times10$}    
     \\\cmidrule(lr){2-4}\cmidrule(lr){5-7}
     & NLL$\downarrow$ & RNLL($\downarrow$) & ESS($\uparrow$) & NLL$\downarrow$ & RNLL($\downarrow$) & ESS($\uparrow$)\\\midrule
     BG (FKL) & 13.40 $\pm$ 0.07 & -9.48 $\pm$ 0.59 & 0.06 $\pm$ 0.01 & 22.01 $\pm$ 0.08 & -17.60 $\pm$ 0.19 & 0.01 $\pm$ 0.00\\
     BG (RKL) & 23.18 $\pm$ 0.19 & -18.75 $\pm$ 0.44 & 0.08 $\pm$ 0.05 & 29.01 $\pm$ 1.01 & -29.32 $\pm$ 0.78 & \textbf{0.70} $\pm$ \textbf{0.04}\\
     FAB & 13.19 $\pm$ 0.21 & -5.14 $\pm$ 0.59 & 0.10 $\pm$ 0.05 & 21.47 $\pm$ 0.22 & -1.05 $\pm$ 0.86 & 0.0001 $\pm$ 0.00\\
     iDEM & 23.76 $\pm$ 1.05 & -23.17 $\pm$ 2.70 & 0.004 $\pm$ 0.004 & 38.13 $\pm$ 2.13 & -25.64 $\pm$ 7.12 & 0.004 $\pm$ 0.004\\
     FUND(FKL) & 12.48 $\pm$ 0.21 & -13.94 $\pm$ 1.52 & 0.14 $\pm$ 0.04 & 19.59 $\pm$ 0.03 & -17.54 $\pm$ 0.83 & 0.14 $\pm$ 0.01\\
    \midrule
    SIMNF & \textbf{12.35} $\pm$\textbf{ 0.06} & -13.21 $\pm$ 1.25 & \textbf{0.33} $\pm$ \textbf{0.03} & \textbf{19.42} $\pm$\textbf{ 0.02} & -15.47 $\pm$ 0.17 &\textbf{ 0.17} $\pm$ \textbf{0.01}\\
    \bottomrule
    \end{tabular}}
    \vspace{-1mm}
    \caption{Results for the Scalar $\phi^4$ distributions. Reported values represent the average and standard error computed over three random seeds for each method.}
    \label{tab:phi4_results}
\end{table*}

\section{Method}
\label{sec:method}
\subsection{Problem Statement}
Let the target distribution over $\mathbf{x}\in\mathbb{R}^d$ be
\begin{equation}
    p(\mathbf{x})=\frac{\tilde{p}(\mathbf{x})}{Z},
    \quad
    \tilde{p}(\mathbf{x})=\exp[-H(\mathbf{x})],
    \quad
    Z=\int_{\mathbb{R}^d}\tilde{p}(\mathbf{x})\,d\mathbf{x}
\end{equation}
where $H(\mathbf{x})$ is the energy (or Hamiltonian) and $Z$ is unknown and intractable. The goal is to learn a generative model $q_\theta(\mathbf{x})$ such that
$q_\theta(\mathbf{x})\approx p(\mathbf{x})$, using only evaluations of
$\tilde{p}(\mathbf{x})$ (or $H(\mathbf{x})$), without target samples during training.

\subsection{Background}
In this section, we review the fundamental principles of normalizing flows and simulated annealing that form the basis of our proposed framework.

\textbf{Normalising flows (NFs)} \cite{NF_survey,papamakarios2021normalizing} model complex distributions by transforming a tractable base distribution through 
a sequence of invertible mapping. Let $\mathbf{z}\sim p_{\mathbf{z}}(\mathbf{z})$ and $\mathbf{x}=f_{\theta}(\mathbf{z})$, where $f_{\theta}$ is a differentiable bijective mapping. The modelled density is given by
\begin{equation}
    q_{\theta}(\mathbf{x})
    =
    p_{\mathbf{z}}\!\left(f_{\theta}^{-1}(\mathbf{x})\right)
    \left|\det J_{f_{\theta}^{-1}}(\mathbf{x})\right|
\label{density_equation}
\end{equation}
The invertible mapping enables direct generation of samples through the forward transformation and explicit likelihood estimation.

\textbf{Simulated annealing (SA Algorithm)} \cite{kirkpatrick1983optimization} facilitates exploration of complex and multimodal distributions by constructing a sequence of annealed distributions, which progressively sharpen. Given a target $p(\mathbf{x})$, define
\begin{equation}
    p_j(\mathbf{x})
    \propto
    p(\mathbf{x})^{\beta_j},
    \qquad
    0<\beta_1<\beta_2<\cdots<\beta_n=1
\label{sa_equ}
\end{equation}
where $\beta\propto T^{-1}$. The resulting sequence is
\begin{equation}
    p_{1}\rightarrow p_2\rightarrow\cdots\rightarrow p_{n-1}
    \rightarrow p_{n}=p
\end{equation}
A small $\beta_1$ corresponds to a high-temperature, smoothed distribution with a flattened energy landscape, facilitating exploration across the state space. As $\beta$ increases, the temperature decreases and the distribution progressively concentrates around high-probability modes, reaching the target at $\beta_n=1$. States obtained at one stage are used to initialise the next, allowing the sampler to gradually cross energy barriers and explore modes that are well separated in the target distribution.

\subsection{Proposed Method}
\label{sec:proposed_method}

A normalizing flow (NF) trained by minimizing the reverse Kullback--Leibler
(KL) divergence, $\mathcal{KL}(q_{\theta}\|p)$, requires only evaluations of
the unnormalized target density and therefore does not require target samples.
However, its mode-seeking behavior can lead to mode collapse for
high dimensional multimodal distributions.

To alleviate this limitation, we incorporate simulated annealing into the NF
framework. Given the target distribution $p(\mathbf{x})$, we construct
\vskip -0.4cm
\begin{equation}
    p_j(\mathbf{x}) \propto p(\mathbf{x})^{\beta_j},
    \qquad
    0<\beta_1<\beta_2<\cdots<\beta_n=1
\end{equation}
Small values of $\beta_j$ correspond to high-temperature, smoothed
distributions with attenuated energy variations, facilitating exploration of
the state space across modes~\cite{kirkpatrick1983optimization,kanaujia2026fund}. The flow is trained sequentially along this annealing path, with each
stage initialized using the parameters learned at the preceding stage.

At each stage, we use a weighted reverse KL (RKL) objective
\begin{align}
    \mathcal{L}(\theta)
    =
    \sum_{j=1}^{n}\lambda_j
    \mathcal{KL}(q_{\theta}\|p_j)
    \label{eq:final_loss}
\end{align}
where \( \lambda_j\geq0,
    \sum_{j=1}^{n}\lambda_j=1\).
    
Since $p_j(\mathbf{x})\propto p(\mathbf{x})^{\beta_j}$, this objective can be
written as
\begin{align}
    \mathcal{L}(\theta)
    &=
    \mathbb{E}_{q_{\theta}}
    \left[
    \log q_{\theta}(\mathbf{x})
    -\beta_{\mathrm{eff}}\log p(\mathbf{x})
    \right]+C
    \label{eq:effective_rkl}
\end{align}
where
\begin{equation}
    \beta_{\mathrm{eff}}
    =
    \sum_{j=1}^{n}\beta_j\lambda_j,
    \qquad
    0<\beta_{\mathrm{eff}}\leq1
\end{equation}
and $C$ is independent of $\theta$. Thus, the weighted objective is
equivalent to an RKL objective with an effective annealing parameter
$\beta_{\mathrm{eff}}$.

Training proceeds over $N$ stages by gradually increasing
$\beta_{\mathrm{eff}}$ from $\beta_1$ to $1$. Initially,
\vspace{-5pt}
\begin{equation}
    \lambda_1=1,\qquad
    \lambda_2=\cdots=\lambda_n=0
\end{equation}
so that $\beta_{\mathrm{eff}}=\beta_1$. At each subsequent stage, the weights
are adjusted to increase $\beta_{\mathrm{eff}}$, while the flow parameters
are initialized from the preceding stage. This progressively transfers the
learned representation toward the target distribution. At the final stage,
$\beta_{\mathrm{eff}}=1$, and the objective reduces to
\vskip -0.4cm
\begin{equation}
    \mathcal{L}(\theta)
    =
    \mathbb{E}_{q_{\theta}}
    \left[
    \log\frac{q_{\theta}(\mathbf{x})}{p(\mathbf{x})}
    \right]
    =
    \mathcal{KL}(q_{\theta}\|p)
\end{equation}
By starting from a smooth target and progressively sharpening it, SIMANF reduces the tendency of reverse KL training to prematurely concentrate on a subset of modes. The weighted objective preserves the contribution of smoother stages, helping the flow retain broader modal structure while progressively refining the target distribution.

\subsubsection{Post Annealing Refinement}

Following annealing, we perform a refinement step that combines forward and
reverse KL objectives. Since target samples are unavailable, the forward KL
is estimated using importance weighting of samples generated from
$q_{\theta}$:
\vskip -0.4cm
\begin{align}
    \mathcal{KL}(p\|q_{\theta})
    &\equiv
    -\mathbb{E}_{\mathbf{x}\sim p}
    [\log q_{\theta}(\mathbf{x})] \\
    &=
    -\mathbb{E}_{\mathbf{x}\sim q_{\theta}}
    [w(\mathbf{x})\log q_{\theta}(\mathbf{x})]
    \label{eq:fkl_importance}
\end{align}
where $w(\mathbf{x})=
    \frac{p(\mathbf{x})}{q_{\theta}(\mathbf{x})}$. Since $p$ is known only up to a normalization constant, normalized
importance weights are used. The refinement objective is
\vskip -0.4cm
\begin{equation}
    \mathcal{L}_{\mathrm{R}}(\theta)
    =
    \alpha_1\mathcal{KL}(p\|q_{\theta})
    +
    \alpha_2\mathcal{KL}(q_{\theta}\|p)
    \label{eq:tuning_loss}
\end{equation}
This refinement combines the mode coverage tendency of forward KL with the
mode seeking behavior of reverse KL to further improve the learned
distribution.

For the Mixture of Gaussians (MOG)-8 distribution, we conduct an experiment to visualize the evolution of generated samples across the training stages in Fig.~\ref{fig1}. The samples progressively evolve from a diffuse distribution into well-separated modes, illustrating how the annealing procedure progressively recovers the multimodal structure of the target distribution.

\section{Experiments}
\label{sec:exp_results}
This section describes the benchmark distributions considered in our experiments and the metrics used to evaluate the performance of the proposed method.

\subsection{Distributions}
\textbf{Many-Well (MW) Distribution.}
The $d$-dimensional MW potential consists of $d/2$ independent copies of a two-dimensional double-well potential~\cite{noe2019boltzmann,midgleyflow}, resulting in $2^{d/2}$ modes:
{\small
\vspace{-6pt}
\begin{equation}
H(\mathbf{x}) =
\sum_{i=1}^{d/2}
\left(
    x_{2i-1}^{4}
    - 6x_{2i-1}^{2}
    - 0.5x_{2i-1}
    + 0.5x_{2i}^{2}
\right),
\mathbf{x}\in\mathbb{R}^{d}
\end{equation}
}
We consider MW-8 and MW-16 corresponding to $d=8$ and $16$ with $16$ and $256$ modes, respectively. Following~\cite{midgleyflow}, we generate $10{,}000$ test samples using rejection sampling. The BG(FKL) baseline additionally uses $10{,}000$ training and $10{,}000$ validation samples; the remaining methods are trained without target samples.

\textbf{Scalar $\phi^4$ Theory Distribution.} The scalar $\phi^4$ theory describes a real scalar field $\mathbf{x}\in\mathbb{R}^d$ on a two-dimensional square lattice with $d$ sites and energy given by
\vskip -0.4cm
\begin{equation}
H(\mathbf{x}) =
\sum_{l=1}^{d}
\left[
\lambda x_l^4 + m^2 x_l^2
+ 2\sum_{l'\in n(l)}
\left(x_l^2-x_lx_{l'}\right)
\right]
\end{equation}
where $n(l)$ denotes the set of nearest neighbours of the $l$-th site. Following~\cite{singha2023conditional,kanaujia2025scorenf}, we set $\lambda=4$ and $m^2=-4$ and consider $8\times8$ ($d=64$) and $10\times10$ ($d=100$) lattices. For evaluation, $10{,}000$ test samples are generated using HMC, while training and validation samples are used only by the BG(FKL) baseline.
\subsection{Metrics}
We evaluate the learned distributions using negative log-likelihood (NLL), reverse negative log-likelihood (RNLL), and effective sample size (ESS).

\textbf{NLL:}
NLL measures the fit of $q_\theta$ to target samples:
\begin{equation}
    \mathrm{NLL}
    =-\frac{1}{N}\sum_{i=1}^{N}\log q_\theta(\mathbf{x}_i),
    \qquad \mathbf{x}_i\sim p
\end{equation}
Lower NLL indicates better agreement with the target distribution.

\textbf{RNLL:}
RNLL \cite{kanaujia2025scorenf} evaluates the target density on samples generated by $q_\theta$:
\begin{equation}
    \mathrm{RNLL}
    =-\mathbb{E}_{\mathbf{x}\sim q_\theta}
    [\log p(\mathbf{x})]
\end{equation}
NLL and RNLL jointly characterize the learned distribution: high NLL with low RNLL suggests mode collapse, whereas low NLL with high RNLL suggests excessive coverage of low density regions. Low values of both indicate good agreement with the target.

\textbf{ESS:}
ESS measures the effective number of independent samples:
\begin{equation}
    \mathrm{ESS}
    =
    \frac{\left(\frac{1}{N}\sum_i w_i\right)^2}
    {\left(\frac{1}{N}\sum_i w_i^2\right)},
    \qquad
    w_i=\frac{p(\mathbf{x}_i)}{q_\theta(\mathbf{x}_i)}
\end{equation}
Higher ESS indicates more efficient sampling.

We compare SIMANF with BG (FKL), BG (RKL), FAB, iDEM, and FUND.

\begin{table}[t]
\vskip -0.4cm
\centering
\caption{Training and inference runtime comparison across various methods for scalar $\phi^4$ distribution $(10\times10)$.}
\label{table_runtime_comparison}
\scalebox{0.90}{
\begin{tabular}{l|p{2.0cm}p{2.0cm}}
\toprule
Method & Training Time & Inference Time  \\
\midrule
RKL & 2.35 hrs & 1.2  sec \\
FAB & 10.20 hrs & 1.2 sec \\
iDEM & 1.33 hrs & 34.0 sec\\
FUND(FKL) & 10.32 hrs & 1.2 sec \\
SIMNF & 7.02 hrs & 1.2 sec\\
\bottomrule
\end{tabular}}
\label{tab:phi4_runtime_results}
\end{table}
\vspace{-10pt}
\begin{table}[t]
\vskip -0.4cm
\centering
\caption{Ablation results for the scalar $\phi^4$ distribution on a 
$10\times10$ lattice size ($d=100$).}
\label{tab:ablation}
\scalebox{0.90}{
\begin{tabular}{lccc}
\toprule
Stage & NLL $\downarrow$ & RNLL $\downarrow$ & ESS $\uparrow$ \\
\midrule
Annealing (Stage I)  & 19.70 & -15.23 & 0.11 \\
Refinement (Stage II) & 19.42 & -15.49 & 0.17 \\
\bottomrule
\end{tabular}
}
\vskip -0.4cm
\end{table}

\section{Results and Discussion}
\label{sec:results}
\subsection{MW-8 and MW-16}
Table~\ref{tab:mw_results} compares SIMANF with the baseline methods on the MW-8 and MW-16 distributions. SIMANF achieves consistently low NLL and high ESS, indicating good agreement with the target and efficient sampling. While BG (RKL) attains the highest ESS, its substantially higher NLL and low RNLL suggest concentration on a limited subset of high density modes, indicating mode collapse. SIMANF achieves the lowest NLL for both distributions, with NLL and ESS comparable to FUND (FKL), and substantially better NLL and ESS than FAB and iDEM. In contrast, BG (FKL) has lower ESS, while its low NLL and high RNLL suggest excessive mode coverage, with probability mass extending into low or near-zero density regions.


\subsection{Scalar $\phi^4$ Distribution}
Table~\ref{tab:phi4_results} compares SIMANF with the baseline methods on the $8\times8$ and $10\times10$ $\phi^4$ theory distributions. For the $8\times8$ lattice, SIMANF achieves the lowest NLL and highest ESS among all methods. Its substantially lower NLL and higher ESS than direct RKL training highlight the effectiveness of the annealing in improving density estimation and sampling efficiency. On the more challenging $10\times10$ lattice distribution, SIMANF again achieves the lowest NLL, with ESS substantially higher than FAB and iDEM and comparable to FUND (FKL). Although BG (RKL) attains a higher ESS, its substantially higher NLL indicates poorer agreement with the target distribution. In contrast, BG (FKL) and FAB, despite achieving competitive NLL values, exhibit very low ESS, indicating that good density fit does not necessarily translate to efficient sampling. iDEM exhibits high NLL and low RNLL, suggesting concentration on a limited subset of high-density modes, while its very low ESS indicates poor sampling efficiency.

In Table~\ref{tab:phi4_runtime_results}, we compare the computational cost of training and generating 10,000 samples for the $10\times10$ lattice on an NVIDIA A100 SXM4 GPU with 40~GB memory. iDEM has relatively low training cost but incurs substantial sample generation overhead due to iterative reverse time SDE simulation. FAB and FUND incur higher training cost because of their iterative procedures and additional computations, including AIS sampling for FAB and multiple annealing stages with a composite objective for FUND. Although SIMANF also uses sequential training, its simpler objective results in lower training cost than FAB and FUND. At inference, SIMANF retains the efficiency of normalizing flows, generating samples with a single forward pass.

We further perform an ablation study on the $10\times10$ scalar $\phi^4$ distribution ($d=100$) to assess the contribution of the post annealing refinement stage. As shown in Table~\ref{tab:ablation}, the refinement yields only a marginal improvement in NLL while substantially improving ESS, indicating enhanced sampling efficiency. These results highlight the complementary role of refinement following annealing.

\section{Conclusions}
\label{sec:conc}
In this work, we present SIMANF, a sample free framework that combines simulated annealing with normalizing flows for learning high-dimensional, multimodal unnormalised distributions. By progressively increasing the annealing parameter and transferring the learned flow between successive stages, SIMANF reduces mode collapse observed with direct reverse KL training while promoting broader mode coverage.  Following annealing, an importance weighted forward KL refinement using flow generated samples further improves density estimation and sampling efficiency. Experiments on Many-Well and high dimensional $\phi^4$ lattice field theory distributions demonstrate consistent performance across increasing dimensions. Although annealing introduces additional training cost, computational analysis shows that SIMANF provides a favourable balance between training cost and sampling efficiency, while retaining efficient inference through a single forward pass. These results highlight annealing based training as a promising approach for sample free learning of complex unnormalised distributions.


---------------------------------



\vfill\pagebreak

\bibliographystyle{IEEEbib}
\bibliography{strings,refs}

\end{document}